\documentclass[twoside,11pt]{article}

\usepackage[preprint]{jmlr2e}

\newcommand\eg{\emph{e.g.,\ }}

\jmlrheading{21}{2021}{1-5}{10/21}{12/21}{tribuo21a}{Adam Pocock}

\ShortHeadings{Tribuo: ML and Provenance in Java}{Pocock}
\firstpageno{1}

\begin{document}

\title{Tribuo: Machine Learning with Provenance in Java}

\author{\name Adam Pocock \email adam.pocock@oracle.com\\
       \addr Machine Learning Research Group\\
       Oracle Labs\\
       Burlington, MA, USA}

\editor{MLOSS editors}

\maketitle

\begin{abstract}%   <- trailing '%' for backward compatibility of .sty file
Machine Learning models are deployed across a wide range of industries,
performing a wide range of tasks.  Tracking these models and ensuring they
behave appropriately is becoming increasingly difficult as the number of
deployed models increases. There are also new regulatory burdens for ML systems
which affect human lives, requiring a link between a model and its training
data in high-risk situations. Current ML monitoring systems often provide
provenance and experiment tracking as a layer on top of an ML library,
allowing room for imperfect tracking and skew between the tracked
object and the metadata. In this paper we introduce Tribuo, a Java ML
library that integrates model training, inference, strong type-safety,
runtime checking, and automatic provenance recording into a single
framework.  All Tribuo's models and evaluations record the full processing
pipeline for input data, along with the training algorithms,
hyperparameters and data transformation steps automatically. The provenance
lives inside the model object and can be persisted separately using common
markup formats.  Tribuo implements many popular ML algorithms for
classification, regression, clustering, multi-label classification and
anomaly detection, along with interfaces to XGBoost, TensorFlow and ONNX
Runtime. Tribuo's source code is available at
\url{https://github.com/oracle/tribuo} under an Apache 2.0 license with
documentation and tutorials available at \url{https://tribuo.org}.
\end{abstract}


\section{Introduction}

Machine Learning models are increasingly deployed across industries to improve
computer systems and integrate with large existing software systems. There has
been an explosion of tools to help companies and individuals train, monitor and
deploy ML models. Several of these tools provide provenance and reproducibility
support to existing libraries (\eg TF Extended \citep{karmarkar2020towards},
Weights \& Biases \citep{wandb}), though most of them provide this tracking by
modifying or parsing existing scripts rather than direct integration with the
libraries performing the ML computations. As a consequence they may not capture
all the interactions between a script and data and may increase the overhead of
building ML systems, both for the data scientist and computationally. In
contrast, Tribuo provides a comfortable type-safe ML
library that integrates \emph{provenance} as a core feature built into
\emph{every} model and evaluation with no additional overhead. To facilitate 
integration with large software systems we've implemented Tribuo in Java, as
the JVM is the most popular platform for building enterprise software \citep{cloudfoundry2018}.

Tribuo's design has four core elements: features and output dimensions are
\emph{named} rather than indexed, feature spaces are sparse by default, objects
in Tribuo are strongly typed with runtime validation checks, and all operations
that result in datasets, models or evaluations are tracked in provenance
objects.  Consequently Tribuo's data points are objects containing sparse
arrays of features, where each feature has a name and a numeric value, along
with an output object which represents the ground truth value for the
prediction task (\eg for classification, a label string; for regression, a
numeric value).  The first two elements emerged from an initial focus on
Natural Language Processing, but have proved widely useful for inference time
feature space validation and working with complex tabular datasets. The latter
two elements follow from our experience deploying ML systems in large
enterprise codebases, where compile time checking is important, and provenance
allows us to trace and debug models in production.

\section{System design}

Machine Learning systems have a variety of constraints that must be satisfied
for reliable deployment of models, however these constraints are often hidden
by the model interface which accepts a multidimensional array as input and
produces a multidimensional array as output. For example, classification models
cannot be directly applied to regression tasks, models must receive features
from the space they have been trained on, many models must be trained before
they can be applied. As machine learning becomes more prevalent, mismatches in
these properties cause issues with deployed ML systems which can be hard to
diagnose such as asking a NLP system to make a prediction when it was not
trained on any of the input words. In Tribuo, we prevent several classes of
user error through a mixture of compile-time and runtime checks. We use Java's
type system to enforce that models are trained before they are applied, and
that classification models are only trained and applied on classification
tasks. At runtime, we check that the features present in the input data have
the correct names, dropping features that the model hasn't seen and raising an
exception if there is no feature overlap between the input data and the model.
We also capture feature metadata that can be used to raise additional warnings
if, for example, the input data is outside the range of the features which the
model was trained on.

The nature of these checks means that Tribuo requires stronger contracts
between its data sources, model trainers and the models themselves. This
contrasts with many ML libraries that follow scikit-learn's
\citep{pedregosa2011scikit} lead in expecting a minimal interface of two
functions (\texttt{fit} and \texttt{predict}) that accept anything 
``array-like''. Tribuo's model trainers (which implement the \texttt{Trainer}
interface) only accept Tribuo's \texttt{Dataset} objects as inputs for model
training. This allows them to enforce that the type of the dataset (\eg
Classification, Regression) matches the type of the trainer and that the
dataset contains the necessary provenance information to describe itself.
Similarly Tribuo's \texttt{Model} objects can only make predictions on examples
that have the correct output type, enforcing that when loading an unknown model
from disk you get an error message rather than silently using a regression
model to make predictions on a classification task. While these type
restrictions, along with the provenance requirements, make it more complicated
to extend Tribuo with new models or data loaders, they provide stronger
guarantees for code built on top of Tribuo and do not increase the complexity
of \emph{using} Tribuo.  We think this trade-off is acceptable as users of
Tribuo outnumber the number of people developing Tribuo itself.

Tribuo has Java implementations of CART \citep{friedman2001elements},
Classifier Chains \citep{read2011classifier}, Factorization Machines
\citep{rendle2010factorization}, Linear Chain Conditional Random Fields
\citep{lafferty2001conditional}, linear \& logistic regressions trained using
SGD \citep{bottou2010large}, multinomial Naive Bayes and sparse linear
regressions using the Lasso and Elastic Net penalties
\citep{friedman2001elements}. There are a variety of SGD algorithms implemented,
including popular ones from the neural network literature such as Adam
\citep{kingma2014adam} and AdaGrad \citep{duchi2011adaptive}, which can be
applied to the factorization machines, CRFs and linear models.  It also has
Java implementations of multi-class Adaboost, Bagging, Random Forests
\citep{friedman2001elements} and extremely randomized trees
\citep{geurts2006extremely}, where Adaboost and Bagging can use any other
trainer as the base learner. In addition to the native Tribuo implementations
it also wraps the Java implementations of LibSVM and LibLinear
\citep{fan2008liblinear} (the latter via the liblinear-java implementation
\citep{liblinear-java}), and it uses the Java interfaces to the C
implementations of TensorFlow \citep{abadi2016tensorflow}, ONNX Runtime
\citep{onnxruntime} and XGBoost \citep{chen2016xgboost}. Tribuo runs on Java 8
and newer versions, while the native libraries (ONNX Runtime, TensorFlow and
XGBoost) are available for x86\_64 platforms running on Windows, macOS and
Linux. ONNX Runtime and XGBoost also work on ARM64 platforms, though users will
need to compile these dependencies themselves. Many Tribuo models can be
exported in the ONNX model format\footnote{\url{https://onnx.ai/}} for
deployment in other languages, on accelerator hardware, and on cloud platforms.

\section{Provenance for data, models and evaluation}

The history or creation path of an object or piece of data is referred to in
the academic literature as its \emph{provenance} or \emph{lineage}.  Tracking
the provenance of Machine Learning datasets, models and evaluations is an
active area of research, which principally focuses on monitoring the behaviour
of the system to record how data flows into a model
\citep{namaki2020vamsa,phani2021lima}, both training data and other information
such as hyperparameter values and algorithm choices. Such systems imply some
overhead in the tracking (\eg Vamsa reparses Python scripts to extract
provenance) and a potential lack of fidelity if the system did not understand
some part of the computation.  Tracking the training data used by any given
model is increasingly important for GDPR compliance and regulations like the
EU's proposed AI regulatory
framework\footnote{\url{https://digital-strategy.ec.europa.eu/en/policies/regulatory-framework-ai}}
which require that all ``high-risk'' models can be traced back to their
training data, meaning that data provenance will likely be further integrated
into Machine Learning Systems.

In Tribuo, we chose to build provenance into the core of the system, each
object (such as a data source, data transformation, training algorithm or
model) knows exactly how it was created and the provenance of all elements that
were used to construct/train it. It is a necessary part of implementing
Tribuo's API that the provenance methods return useful values, and code which
does not appropriately use the provenance does not pass our continuous
integration tests. This means that when working inside Tribuo, everything the
user does is tracked and transcribed into the provenance objects automatically,
and those objects are themselves fields of the host object that is the subject
of the provenance. As a result, the Tribuo provenance object mirrors perfectly
the host object and is low overhead (creating small Java objects is fast and
cheap). Tribuo's provenance can be conceptually split into two kinds,
configuration and instance information. Configuration describes
information statically known before the computation begins, \eg model
hyper-parameters, the path to training data files, training algorithm choice.
Instance information is derived from the computation as it is executed, \eg
SHA-256 hashes of the training data, user supplied information, OS and machine
architecture. The configuration aspects can be separated out of a provenance
object and stored separately as a configuration file in a variety of formats
(currently xml, json, protocol buffers and edn) for running reproducible
experiments.

The central provenance object is \texttt{ModelProvenance}. It contains a
\texttt{DataProvenance}, which is extracted from the training dataset, and a
\texttt{TrainerProvenance}, which is extracted from the training algorithm.  If
the model is an ensemble then it also contains a list of
\texttt{ModelProvenance} objects, one per ensemble member. The
\texttt{DataProvenance} object tracks the number of features, the number of
examples, any transformations applied both at a per feature and global level
(e.g., zero mean unit variance rescaling), and the original
\texttt{DataSourceProvenance}. The \texttt{DataSourceProvenance} tracks the
location the data was loaded from, whether a file on disk, a DB connection, or
created in memory, along with a hash of the data and other relevant information
such as the feature extraction procedure and the mapping between columns and
features. The \texttt{TrainerProvenance} includes the training algorithm (along
with any nested algorithms in the case of an ensemble), the algorithm's
hyperparameters (\eg learning rate, tree depth), and any RNG seeds to ensure
reproducibility. When evaluating models, the \texttt{ModelProvenance} is stored
inside the evaluation alongside the \texttt{DataProvenance} for the test data.
To prevent the provenance from becoming stale and to avoid having to figure out
which provenance applies to which model, we store these provenance objects
\emph{inside the model object}.  If necessary to maintain confidentiality the
provenance can be redacted and replaced with a hash which can be linked to the
provenance in an external database, similar to other model tracking systems.

\section{Uses of model and data provenance}

We have two main uses for provenance information in Tribuo's models and
evaluations.  The first is simply tracking models via their metadata. As the
provenance is stored in the model object both in memory and on disk, it is
guaranteed to be present with the model and so is hard to confuse with the
provenance for a different model. When there are hundreds of models in
production, this makes it simple to understand what data a specific model was
trained on and what the training algorithm was without resorting to a
potentially inaccurate external system. The models are \emph{self-describing},
they know properties of their training data like the input feature
distributions, the output label distributions and the number of training
examples, in addition to a full description of the training algorithm used,
along with any feature transformations applied to the data before training.

The second use is as a way of storing the input pipeline in a recoverable
fashion inside the model object. This is most useful when using Tribuo's
columnar data system to featurise inputs. The columnar package can extract
features of different kinds from input strings, either by converting them
directly into numeric feature values, binarising categorical variables, or
applying a full text processing pipeline among other options. The processing
infrastructure has its provenance recorded as part of the training data
provenance, stored inside the model object.  Using the configuration extractor
the columnar provenance can be converted into configuration for the columnar
processor and the processing object can be re-instantiated, ready to process
new inputs after the model has been loaded.  This system is not specific to
Tribuo's columnar processor, \emph{any} Tribuo data loader can be reconstructed
from the provenance, and the system is not closed world, it can be extended by
implementing the appropriate interfaces in user classes outside of Tribuo's
namespace.

Over time we plan to expand Tribuo's use of provenance information, first to
add full automatic reproducibility of Tribuo trained models and evaluations and
then to use the reproducibility framework as a basis for an experimental
tracking system that integrates hyperparameter optimization.

\section{Conclusion}

In this paper we presented Tribuo, a ML library for the Java platform, that has
a strong focus on type-safety, runtime checks, and metadata tracking via
provenance objects. The focus on provenance makes Tribuo well suited for use in
systems where any data and algorithms used in model creation must be tracked
for compliance or regulatory reasons. Tribuo's source code is available at
\url{https://github.com/oracle/tribuo} under an Apache 2.0 license with
documentation and tutorials available at \url{https://tribuo.org}.

% Acknowledgements
\acks{We would like to acknowledge the contributions of the Machine Learning
Research Group at Oracle Labs, along with our internal collaborators at Oracle,
and our external contributors in the open source community.}

\vskip 0.2in
\bibliography{tribuo-jmlr}

\end{document}